\documentclass[10pt,twocolumn,letterpaper]{article}

\usepackage[pagenumbers]{cvpr} % To force page numbers, e.g. for an arXiv version
\usepackage[accsupp]{axessibility}

\usepackage{graphicx}
\usepackage{amsmath}
\usepackage{amssymb}
\usepackage{booktabs}
\usepackage{multirow}
\usepackage{makecell}
\usepackage{pifont}
\usepackage{verbatim}
\usepackage{booktabs,caption}
\usepackage[flushleft]{threeparttable}

\newcommand{\cmark}{\ding{51}}%
\newcommand{\xmark}{\ding{55}}%

\usepackage[pagebackref,breaklinks,colorlinks]{hyperref}

\usepackage[capitalize]{cleveref}
\crefname{section}{Sec.}{Secs.}
\Crefname{section}{Section}{Sections}
\Crefname{table}{Table}{Tables}
\crefname{table}{Tab.}{Tabs.}

\def\cvprPaperID{3688} % *** Enter the CVPR Paper ID here
\def\confName{CVPR}
\def\confYear{2022}
\def\Ours{{Focals Conv}\xspace}
\def\OursF{{Focals Conv-F}\xspace}

\begin{document}

\title{Focal Sparse Convolutional Networks for 3D Object Detection}

\author{
Yukang Chen$^{1}$,
~~~
Yanwei Li$^{1}$,~~~
Xiangyu Zhang$^{2}$,~~~
Jian Sun$^2$,~~~
Jiaya Jia$^{1,3}$
\\[0.2cm]
$^1$The Chinese University of Hong Kong~~
$^2$MEGVII Technology~~
$^3$SmartMore
}
\maketitle

%%%%%%%%% ABSTRACT
\begin{abstract}
    
    Non-uniformed 3D sparse data, {\em e.g.}, point clouds or voxels in different spatial positions, make contribution to the task of 3D object detection in different ways. 
    Existing basic components in sparse convolutional networks (Sparse CNNs) process all sparse data, regardless of regular or submanifold sparse convolution. In this paper, we introduce two new modules to enhance the capability of Sparse CNNs, both are based on making feature sparsity learnable with position-wise importance prediction. They are focal sparse convolution (Focals Conv) and its multi-modal variant of focal sparse convolution with fusion, or Focals Conv-F for short. The new modules can readily substitute their plain counterparts in existing Sparse CNNs and be jointly trained in an end-to-end fashion. For the first time, we show that spatially learnable sparsity in sparse convolution is essential for sophisticated 3D object detection. Extensive experiments on the KITTI, nuScenes and Waymo benchmarks validate the effectiveness of our approach. Without bells and whistles, our results outperform all existing single-model entries on the nuScenes test benchmark. 
    Code and models are at \href{https://github.com/dvlab-research/FocalsConv}{github.com/dvlab-research/FocalsConv}.
    \footnotetext{Yukang's work was done during internship in MEGVII Technology.}
    
\end{abstract}

%%%%%%%%% BODY TEXT
\section{Introduction}
\label{sec:intro}
% first paragraph
% Key challenge for 3D object detection -- unstructured data
% Voxel-based methods (Sparse, outdoor, autonomous driving), Point based methods.
A key challenge in 3D object detection is to learn effective representations from the unstructured and sparse 3D geometric data such as point clouds. In general, there are two ways for this job. The first is to process point clouds~\cite{point-rcnn, 3dssd} directly, based on PointNet++~\cite{pointnet++} networks. However, the neighbour sampling and grouping operations are time-consuming. This makes it improper for large-scale autonomous driving scenes that require real-time efficiency. The The second is to convert point clouds into voxelizations and apply 3D sparse convolutional neural networks~(Sparse CNNs) for feature extraction~\cite{pvrcnn,voxel-rcnn}. 3D Sparse CNNs resemble 2D CNNs in structures, including several feature stages and down-sampling operations. They typically consist of {\em regular} and {\em submanifold} sparse convolutions~\cite{submanifold-sparse-conv-v2}.

% second paragraph
% Drawbacks in the current sparse convolution
Although regular and submanifold sparse convolutions have been widely used, they have respective limitations. 
Regular sparse convolution dilates all sparse features. It inevitably burdens models with considerable computations. That is why backbone networks commonly limit its usage only in down-sampling layers~\cite{second,pvrcnn}. In addition, detectors aim to distinguish target objects from massive background features. But regular sparse convolution reduces the sparsity sharply and blurs feature distinctions. 

\begin{figure*}[t]
\begin{center}
   \includegraphics[width=\linewidth]{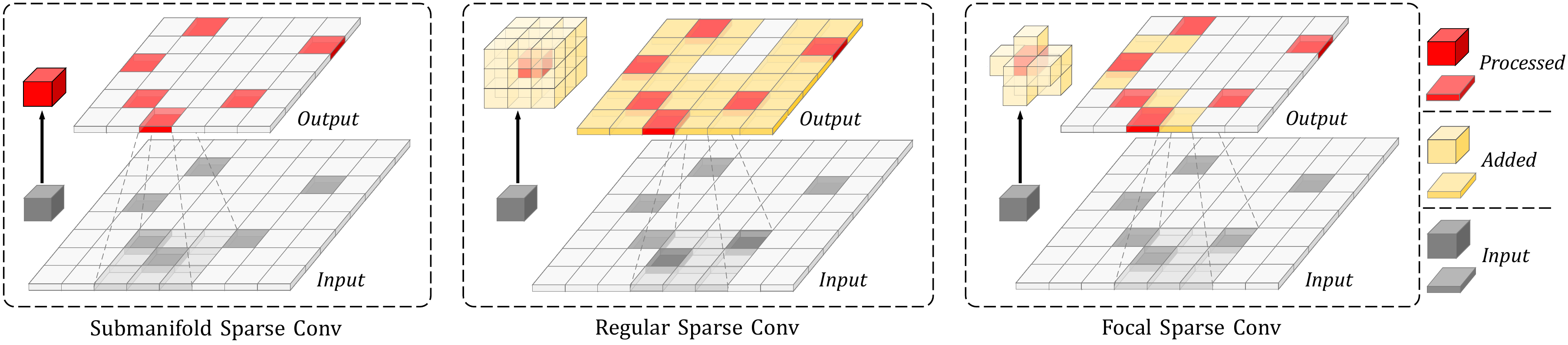}
   \caption{Process of different sparse convolution types. {\em Submanifold sparse convolution} fixes the output position identical to input. It maintains efficiency but disables information flow between disconnected features. {\em Regular sparse convolution} dilates all input features to its kernel-size neighbors. It encourages information communication with expensive computation, as it seriously increases feature density. The proposed {\em focal sparse convolution} dynamically determines which input features deserve dilation and dynamic output shapes, using predicted {\em cubic importance}. {\em Input} and {\em Output} are illustrated in 2D features for simplification. This figure is best viewed in color.}
   \label{fig:illustrate23d}
\end{center}
\end{figure*}

On the other hand, submanifold sparse convolutions avoid the computation issue by restricting the output feature positions to the input. But it misses necessary information flow, especially for the spatially disconnected features. The above issues on regular and submanifold sparse convolutions limit Sparse CNNs to achieve high representation capability and efficiency. We illustrate the submanifold and regular sparse convolutional operations in Fig.~\ref{fig:illustrate23d}.

% third paragraph
% Analyze the reasons behind these drawbacks. the design of the operations influenced by 2D CNN, whose inputs are structured.
These limitations originate from the conventional convolution pattern: all input features are treated equally in the convolution process. It is natural for 2D CNNs, and yet is improper for 3D sparse features. 2D convolution is designed for structured data. All pixels in the same layer typically share receptive field sizes. But 3D sparse data is with varying sparsity and importance in space. It is not optimal to handle non-uniform data with uniform treatment. In terms of {\em sparsity}, upon the distance to LIDAR sensors, objects present large sparsity variance. 
In terms of {\em importance}, the contribution of features varies with different locations for 3D object detection, {\em e.g.}, foreground or background. 
Although 3D object detection is achieved~\cite{point-rcnn, pvrcnn, voxel-rcnn, centerpoint}, state-of-the-art methods still rely on RoI (region-of-interest) feature extraction. It corresponds to the idea that we should shoot arrows at the target in the feature extraction of 3D detectors.

% fourth paragraph
% In this paper, we propose our Focal Sparse Convolution
% Multi-modal focal sparse conv, why it is feasible in image-involved methods.
In this paper, we propose a general format of sparse convolution by relaxing the conceptual difference between regular and submanifold ones. We introduce two new modules that improve the representation capacity of Sparse CNNs for 3D object detection. 
The first is focal sparse convolution~({\em \Ours}). It predicts {\em cubic importance} maps for the output pattern of convolutions. Features predicted as {\em important} ones are dilated into a {\em deformable} output shape, as shown in Fig~\ref{fig:illustrate23d}. The importance is learned via an additional convolutional layer, dynamically conditioned on the input features. This module increases the ratio of valuable information among all features. 
The second is its multi-modal improved version of Focal sparse Convolution with Fusion~(named as {\em \OursF}). Upon the LIDAR-only {\em \Ours}, we enhance importance prediction with RGB features fused, as image features typically contain rich appearance information and large receptive fields.

\begin{figure*}[t]
\begin{center}
   \includegraphics[width=\linewidth]{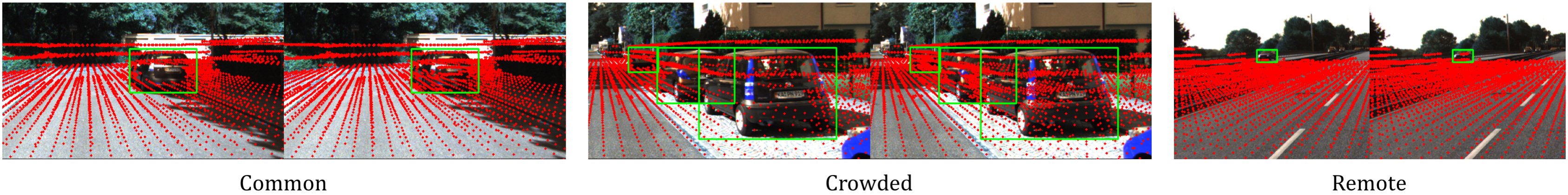}
   \caption{Illustrations on learnable feature sparsity. We project the 3D voxel centers from the backbone output onto 2D image planes. These cases include common, crowded, and remote objects. Left: plain Sparse CNN. Right: focal sparse CNN. Focal sparse convolution adaptively densifies object features without introducing redundant background features. This figure is best viewed in color and by zoom-in.
   %More illustrations are in the {\em supplementary material.
   }
   \label{fig:sparsity_comparison_3pairs}
\end{center}
\end{figure*}

% fifth paragraph
% Attributes of them proposed modules
The proposed modules are novel in two aspects. 
First, {\em \Ours} presents a dynamic mechanism for learning spatial sparsity of features. It makes the learning process concentrated on the more valuable foreground data. With the down-sampling operations, valuable information increases in stages. Meanwhile, the large amount of background voxels are removed. Fig.~\ref{fig:sparsity_comparison_3pairs} illustrates the learnable feature sparsity, including the common, crowded, and remote objects, where {\em \Ours} enriches the learned voxel features on the foreground without redundant voxels added in other areas.
Second, both modules are lightweight. The importance prediction involves small overhead parameters and computation, as measured in Tab.~\ref{tab:improvements-pvrcnn}. 
The RGB feature extraction of {\em \OursF} involves only {\em several layers}, instead of heady 2D detection or segmentation models~\cite{pointpainting}. 

% sixth paragraph
% Introduce the experimental results.

The proposed modules of {\em \Ours} and {\em \OursF} can readily replace their original counterparts in sparse CNNs. To demonstrate the effectiveness, we build the backbone networks on existing 3D object detection frameworks~\cite{pvrcnn, voxel-rcnn, centerpoint}. Our method enables non-trivial enhancement with small model complexity overhead on both the KITTI~\cite{kitti} and nuScenes~\cite{nuscenes} benchmarks. These results manifest that learnable sparsity with focal points is essential. 
Without bells and whistles, our approach outperforms state-of-the-art ones on the nuScenes {\em test} split~\cite{nuscenes}. 

% sixth paragraph
% sumarize the contributions.
Convolutional dynamic mechanism adapts the operations conditioned on input data, {\em e.g.}, deformable convolutions~\cite{deformableconv, deformableconvv2} and dynamic convolutions~\cite{condconv, dynamicconv}. %They all adapt the convolution operation to be conditioned on input data. 
The key difference is that our approach makes use of the {\em intrinsic sparsity} of data. It promotes feature learning to be concentrated on more valuable information.
We deem the non-uniform property as a great benefit.  We discuss the relations and differences to previous literature in Sec.~\ref{sec:related_work}.

%------------------------------------------------------------------------
\section{Related Work}
\label{sec:related_work}
\subsection{Convolutional Dynamic Mechanism}
Dynamic mechanisms have been widely studied in CNNs, due to their advantages of high accuracy and easy adaption in scenarios. We discuss two kinds of related methods, i.e., %kernel weight attention~\cite{condconv, dynamicconv, pixel-adaptive-conv},
kernel shape adaption~\cite{deformableconv, deformableconvv2, kpconv}, and input attention mask~\cite{dynamicconv-fasterinference, spatialsampling, sbnet}.

\vspace{0.5em}
\noindent
\textbf{Kernel shape adaption.}
Kernel shape adaption methods~\cite{deformableconv, deformableconvv2, deformablekernels,minkowskinet} adapt the effective receptive fields of networks. Deformable convolution~\cite{deformableconv} predicts offsets for feature sampling. Its variant~\cite{deformableconvv2} introduces an additional attention mask to modulate features. %Deformable kernel~\cite{deformablekernels} shows better robustness to geometric transformation. 
For 3D feature learning, KPConv~\cite{kpconv} learns local offsets for kernel points. MinkowskiNet~\cite{minkowskinet} generalizes sparse convolution to arbitrary kernel shape. Overall, these methods modify the input feature sampling process. 

Deformable PV-RCNN~\cite{deformable-pvrcnn} applies offset prediction for feature sampling in 3D object detection. In contrast, focal sparse convolution improves the {output} feature spatial sparsity and makes it learned, helpful for 3D object detection. 

\vspace{0.5em}
\noindent
\textbf{Attention mask on input.}
Methods of~\cite{spatialsampling, dynamicconv-fasterinference, pac, eq-paradigm} seek spatial-wise sparsity for efficient inference. These methods receive dense images and prune unimportant pixels based on attention masks. 
These methods aim to sparsify dense data while we make use of {intrinsic data sparsity}. %PAC~\cite{pac} targets at adaptive kernel weights instead of pruning.
Although SBNet~\cite{sbnet} also utilizes the sparse property, it limits application to 2D BEV (bird-eye-views) images, and shares the static masks over all layers in the network. In contrast, our improved convolution is more adaptive and is applicable to related tasks, e.g., 3D instance segmentation~\cite{icm-3d}.

\begin{figure*}[t]
\begin{center}
   \includegraphics[width=\linewidth]{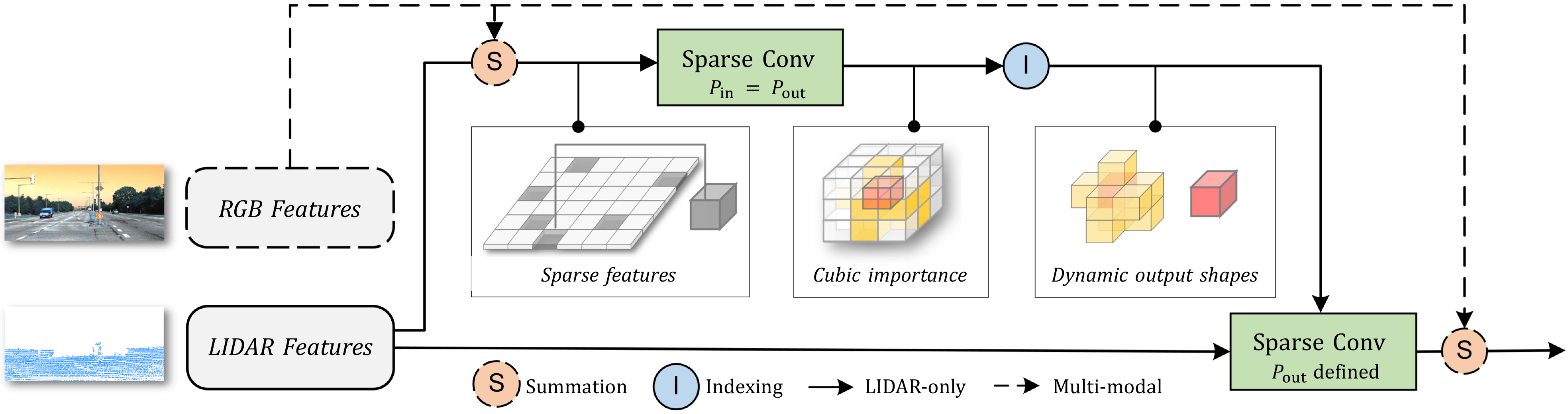}
   \caption{Framework of focal sparse convolution and its multi-modal extension. An additional branch predicts a cubic importance map for each input sparse feature, which determines the output feature positions. In the multi-modal version, the additional branch takes fusion of LIDAR and RGB features for better prediction. Output sparse features predicted as important are also fused with the RGB features.}
   \label{fig:pipeline}
\end{center}
\end{figure*}

% 3D Object Detection

\subsection{3D Object Detection}
% Point Cloud
\noindent
\textbf{LIDAR-only detectors.}
%3D object detection~\cite{second, pointpillars, votenet, pvrcnn} aims to predict 3D rotated bounding boxes and corresponding categories. 
3D object detection frameworks usually resemble 2D detectors, {\em e.g.}, the R-CNN family~\cite{point-rcnn, voxel-rcnn, pvrcnn, pyramid-rcnn} and the SSD family~\cite{3dssd, sessd, sassd, cia-ssd}. 
The main difference on 2D detectors lies in input encoders.
VoxelNet~\cite{voxelnet} encodes voxel features using PointNet~\cite{pointnet} and applies a RPN~(region proposal network)~\cite{fasterrcnn}. SECOND~\cite{second} uses accelerated sparse convolutions and improves efficiency from VoxelNet~\cite{voxelnet}. VoTr~\cite{voxeltransformer} applies transformer architectures to voxels. Various detectors~\cite{pvrcnn, voxel-rcnn, centerpoint} have been presented based on feature encoders. We validate the proposed approach on backbones of frameworks of~\cite{pvrcnn,voxel-rcnn,centerpoint} on multiple datasets~\cite{kitti,nuscenes,waymo}.

\vspace{0.5em}
\noindent
\textbf{Completion-based detectors.}
% Point Completion
Completion-based methods~\cite{pcrgnn, sienet, spg, gsdn} form another line of efforts in enriching foreground information. We focus on feature learning instead of point completion.
PC-RGNN~\cite{pcrgnn} has a point completion module by a graph neural network. SIENet~\cite{sienet} builds upon PCN~\cite{pcn} for point completion in a two-stage framework. The completion process relies on the prior generated proposals. GSDN~\cite{gsdn} expands all features first through transposed convolutions and then by pruning. SPG~\cite{spg} designs a semantic point generation module for domain adaption 3D object detection. It is applied during data preprocessing, complicating the detection pipelines. 
%In contrast, our end-to-end approach makes effects in the input encoding process and is more economical. 

% Cross Modal
\vspace{0.5em}
\noindent
\textbf{Multi modal fusion.}
%Representative multi-modal fusion methods~\cite{3dcvf, deepfusion, pointaugmenting, pointpainting, mutliview3ddet} in 3D object detection can roughly be categorized in to region-level~\cite{mutliview3ddet, joint3dproposal} and pixel-level fusion methods~\cite{mvxnet, epnet}.
Multi-modal fusion methods~\cite{3dcvf, deepfusion, epnet} use more information than LIDAR-only ones.
The KITTI~\cite{kitti} benchmark had been dominated by LIDAR-only methods until PointPainting~\cite{pointpainting} was proposed. It decorates raw point clouds with the corresponding image segmentation scores. PointAugmenting~\cite{pointaugmenting} further replaces the segmentation model with an 2D object detection one~\cite{centernet}. They are both decoration-based methods, which require image feature extraction on off-the-shelf 2D networks, before feeding into 3D detectors. Although promising results are achieved by these methods, the overall inference pipelines are complicated.
Our multi-modal focal sparse convolution differs from the above methods in two aspects. First, we only require several jointly trained layers for image feature extraction, rather than the heavy segmentation or detection models. Second, we only strengthen the predicted {\em important} features, instead of the uniform decoration~\cite{pointpainting,pointaugmenting} for all LIDAR features.

%------------------------------------------------------------------------
\section{Focal Sparse Convolutional Networks}
\label{sec:focal-sparse-convnet}

In this section, we first review the formulation of sparse convolution in Sec.~\ref{sec:review-sparse-conv}. Then, the proposed focal sparse convolution and its multi-modal extension will be elaborated in Sec.~\ref{sec:focal-sparse-conv} and Sec.~\ref{sec:multi-modal-extension}. We finally introduce the resulting focal sparse convolutional networks in Sec.~\ref{sec:convnetwork}.

\subsection{Review of Sparse Convolution}
\label{sec:review-sparse-conv}
Given an input feature $\mathrm{x}_p$ with number of $c_{\mathrm{in}}$ channels at position $p$ in the $d$ dimensional spatial space, we process this feature by a convolution with kernel weights $\mathrm{w}\in \mathbb{R}^{\mathnormal{K}^d\times c_{\mathrm{in}} \times c_{\mathrm{out}}}$. For example, in the 3D coordinate space, $\mathrm{w}$ contains $c_{\mathrm{in}} \times c_{\mathrm{out}}$ spatial kernels with size 3 and $|\mathnormal{K}^d|=3^3$. The convolution process is represented as
\begin{equation}
    \mathrm{y}_p=\sum_{k\in \mathnormal{K}^d} \mathrm{w}_k \cdot \mathrm{x}_{\bar{p}_k},
\end{equation}
where $k$ enumerates all discrete locations in the kernel space $\mathnormal{K}^d$. $\bar{p}_k=p+k$ is the corresponding location around center $p$, where $k$ is an offset distance from $p$.

This formulation accommodates most types of convolutions with simple modifications. When $p\in \mathbb{Z}$, the common convolution for dense input data is yielded. When $\bar{p}_k$ is added with a learned offset $\Delta \bar{p}_k$, it includes the kernel shape adaption methods, {\em e.g.}, deformable convolutions~\cite{deformableconv, deformableconvv2}. 
Further, if $\mathrm{W}$ equals to a weighted sum $\sum\limits\alpha_i \mathrm{W}^i$, it generalizes to weight attention, {\em e.g.}, dynamic convolution~\cite{condconv,dynamicconv}. Finally, when attention masks are multiplied to the input feature map $\mathrm{x}$, this formulation makes input attention mask methods~\cite{sbnet, spatialsampling}.

For sparse input data, the feature position $p$ does not belong to the dense discrete space $\mathbb{Z}$. The input and output feature spatial space is relaxed to $P_{\mathrm{in}}$ and $P_{\mathrm{out}}$, respectively. The formulation is converted to
\begin{equation}
    \mathrm{y}_{p\in P_{\mathrm{out}}}=\sum_{k\in \mathnormal{K}^d(p, P_{\mathrm{in}})} \mathrm{w}_k \cdot \mathrm{x}_{\bar{p}_k},
    \label{eq:sparse-conv}
\end{equation}
where $\mathnormal{K}^d(p, P_{\mathrm{in}})$ is a subset of $\mathnormal{K}^d$, leaving out the empty position. It is conditioned on the position $p$ and input feature space $P_{\mathrm{in}}$ as
\begin{equation}
    \mathnormal{K}^d(p, P_{\mathrm{in}})=\{k\,|\, p+k\in P_{\mathrm{in}},k\in\mathnormal{K}^d\}.
\end{equation}
If $P_{\mathrm{out}}$ includes a union of all dilated positions around $P_{\mathrm{in}}$ within $\mathnormal{K}^d$ neighbours, this process is formulated as 
\begin{equation}
    P_{\mathrm{out}}=\bigcup_{p\in P_{\mathrm{in}}} P(p, \mathnormal{K}^d),
\end{equation}
where
\begin{equation}
    P(p,{\mathnormal{K}^d})=\{p+k\,|\,k\in\mathnormal{K}^d\}.
    \label{eq:output-positions}
\end{equation}
On this condition, the formulation becomes regular sparse convolution. It acts at all positions where any voxels exist in its kernel space. It does not skip any information gathering in the total spatial space. 

This strategy involves two drawbacks. (\romannumeral1) It introduces considerable computation cost. The number of sparse features is doubled or even tripled, increasing burden for following layers. (\romannumeral2) We empirically find that continuously increasing the number of sparse features may harm 3D object detection~(Tab.~\ref{tab:ablation-dynamicoutput}). Crowded and unpromising candidate features may blur the valuable information. It degrades foreground features and further declines the feature discrimination capacity of 3D object detectors. 

When $P_{\mathrm{in}}=P_{\mathrm{out}}$, submanifold sparse convolution~\cite{submanifold-sparse-conv-v2} is yielded. It happens only when the kernel centers locate at the input, restricting the active positions to input sets. This setting avoids the computation burden, but abandons necessary information flow between disconnected features. Note that the flow is common in the irregular point cloud data. Thus, effective receptive field sizes are constrained by the feature disconnection, which degrade the model capability.

\subsection{Focal Sparse Convolution}
\label{sec:focal-sparse-conv}
% formulation
Regardless of regular or submanifold sparse convolution, output positions $P_{\mathrm{out}}$ are static across all $p\in P_{\mathrm{in}}$, which is undesirable.
In contrast, we perform adaptive determination of sparsity or receptive field sizes in a fine-grained manner. 
We relax output positions $P_{\mathrm{out}}$ to be dynamically determined by the sparse features. We illustrate this proposed process in Fig.~\ref{fig:pipeline} (via solid lines).

In our formulation, output positions $P_{\mathrm{out}}$ generalize to a union of all important positions with their dilated area and other unimportant positions. The dilated areas are deformable and dynamic to input positions. Eq.~\eqref{eq:output-positions} becomes
\begin{equation}
    P_{\mathrm{out}} = \left(\bigcup_{p\in P_{\mathrm{im}}} P(p,{\mathnormal{K}_{\mathrm{im}}^d(p)})\right)\cup P_{\mathrm{in}/\mathrm{im}}.
\end{equation}
We factorize this process into three steps: (\romannumeral1) {\em cubic importance prediction}, (\romannumeral2) {\em important input selection}, and (\romannumeral3) {\em dynamic output shape generation}. 

\vspace{0.5em}
\noindent
\textbf{Cubic importance prediction.}
A cubic importance map $I^p$ involves importance for candidate output features around the input feature at position $p$. Each cubic importance map shares the same shape $\mathnormal{K}^d$ with the main processing convolution kernel weight, $e.g.$, $k^3=3\times3\times3$ with the kernel size 3. It is predicted by an additional submanifold sparse convolution with a sigmoid function. The latter steps depend on the predicted cubic importance maps.

\vspace{0.5em}
\noindent
\textbf{Important input selection.}
In Eq.~\eqref{eq:output-positions}, $P_{\mathrm{im}}$ is a subset of $P_{\mathrm{in}}$. It contains the positions of relatively important input features. We select $P_{\mathrm{im}}$ as
\begin{equation}
    P_{\mathrm{im}} = \{p\,|\, I^p_{0}\geq\tau,p\in P_{\mathrm{in}}\},
    \label{eq:p_im}
\end{equation}
where $I^p_{0}$ is the {\em center} of the cubic importance map at position $p$. And $\tau$ is a pre-defined threshold (Tab.~\ref{tab:ablation-importancesampling} and \ref{tab:importance-threshold}). Our formulation becomes the regular or submanifold sparse convolution when $\tau$ is 0 or 1 respectively. We also find that using top-k ratio to select is an alternative of threshold.

% importance maps
\vspace{0.5em}
\noindent
\textbf{Dynamic output shape generation.}
Features in $P_{\mathrm{im}}$ is dilated to a dynamic shape. The output around $p$ is determined by the dynamic output shape $\mathnormal{K}_{\mathrm{im}}^d(p)$. Note that our deformable output shapes are pruned inside the original dilation without offsets. It is computed similarly to Eq.~\eqref{eq:p_im} as
\begin{equation}
    \mathnormal{K}_{\mathrm{im}}^d(p) = \{k\,|\, p+k\in P_{\mathrm{in}}, I^p_{k}\geq\tau, k\in\mathnormal{K}^d\}.
\end{equation}
We analyze the dynamic output shape in Tab.~\ref{tab:ablation-dynamicoutput}.
For the remaining unimportant features, their output positions are fixed as input, {\em i.e.}, submanifold. We found that directly removing them or using a fully dynamic manner without preserving them makes the training process unstable.

% supervision
\vspace{0.5em}
\noindent
\textbf{Supervision manners.}
\label{par:supervision-manner}
In 3D object detection, we have a prior knowledge that foreground objects are more valuable information. Based on this prior, we apply focal loss~\cite{focalloss} as an objective loss function to supervise the importance prediction. We construct the objective targets for the centers of feature voxels inside 3D ground-truth boxes. We keep its loss weight as 1 for the generality of our modules.

Additional supervision comes from multiplying the predicted cubic importance maps to output features as attention. It makes the importance prediction branch differentiable naturally. It shares motivation with the kernel weight sparsification methods~\cite{sparsecnn} in the area of model compression.
We empirically show that this attention manner benefits the performance for minor classes, {\em e.g.}, Pedestrian and Cyclist on KITTI (investigated in Tab.~\ref{tab:ablation-objectiveloss}).

\subsection{Fusion Focal Sparse Convolution}
\label{sec:multi-modal-extension}
We provide a multi-modal version of focal sparse convolution, as illustrated in Fig.~\ref{fig:pipeline} (via dashed lines). This extension is conceptually simple but effective. We extract RGB features from images and align LIDAR features to them. The extracted features are fused to input and {\em important} output sparse features in focal sparse convolution.

% RGB feature extraction
\vspace{0.5em}
\noindent
\textbf{Feature extraction.}
The fusion module is lightweight. It contains a conv-bn-relu layer and a max-pooling layer. It down-samples the input image to 1/4 resolutions. It is followed by {\em 3 conv-bn-relu layers} with residual connection~\cite{resnet}. The channel number is then reduced to be consistent with that of sparse features, with an MLP layer. This facilitates a simple summation of multi-modal features.

% Sparse feature alignment
\vspace{0.5em}
\noindent
\textbf{Feature alignment.}
A common issue during fusion is misalignment in the 3D-to-2D projection. Point cloud data is commonly processed by transformation and augmentation. Transformations include flip, re-scale, rotation, translation. The typical augmentation is ground-truth sampling~\cite{second}, copying paste objects from other scenes. %These introduce a misalignment along 3D to 2D projection. 
For these invertible transformations, we reverse the coordinates of sparse features with the recorded transformation parameters~\cite{pointaugmenting,moca}. For ground-truth sampling, we copy the corresponding 2D objects onto images. Rather than using an additional segmentation model or mask annotations~\cite{moca}, we directly crop objects in bounding boxes for simplification.

% Multi modal feature fusion
\vspace{0.5em}
\noindent
\textbf{Fusion manners.}
The aligned RGB features are directly fused to sparse features in {\em summation}, as they share the same channel numbers. Although other fusion methods, $e.g.$, concatenation or cross-attention, can be used, we choose the most concise summation for efficiency. The aligned RGB features are fused with sparse features twice in this module. It is first fused to input features for cubic importance prediction. Then we fuse RGB features only to {\em important} output sparse features, {\em i.e.}, the first part in Eq.~\eqref{eq:output-positions}, instead of all of them~(investigated in Tab.~\ref{tab:improvements-multimodal-stage}).

Overall, the multi-modal layers are lightweight in terms of parameters and fusion strtegies. They are jointly trained with detectors. It provides an efficient and economical solution for the fusion module in 3D object detection.

\subsection{Focal Sparse Convolutional Networks}
\label{sec:convnetwork}
Both focal sparse convolution and its multi-modal extension can readily replace their counterparts in the backbone networks of 3D detectors. During training, we do not use any special initialization or learning rate settings for the introduced modules. The importance prediction branch is trained via back-propagation through the attention multiplication and objective loss function as introduced in Sec.~\ref{par:supervision-manner}.

The backbone networks in 3D object detectors~\cite{pvrcnn,voxel-rcnn,centerpoint} typically consist of one stem layer and 4 stages. Each stage, except the first one, includes a regular sparse convolution with down-sampling and two submanifold blocks. In the first stage, there are one~\cite{pvrcnn,voxel-rcnn} or two~\cite{centerpoint} sparse convolutional layers. By default, each sparse convolution is followed by batch normalization~\cite{batchnorm} and ReLU activation. 

We validate focal sparse convolution on the backbone networks of existing 3D detectors~\cite{pvrcnn, voxel-rcnn, centerpoint}. 
We directly apply focal sparse convolution at the last layer of certain stages. We analyze the stages for using our focal sparse convolution in experiments~(ablated in Tab.~\ref{tab:ablation-usingstages} and \ref{tab:improvements-multimodal-stage}).

%------------------------------------------------------------------------
\section{Experiments}
\label{sec:experiments}
We conduct ablations and comparisons for {\em \Ours} and its multi-modal variant. More experiments, such as results on Waymo~\cite{waymo}, are in the {supplementary material}.
\subsection{Setup and Implementation}
\noindent
\textbf{KITTI.}
The KITTI dataset~\cite{kitti} consists of 7,481 samples and 7,518 testing samples. The training samples are split into a {\em train} set with 3,717 samples and a {\em val} set with 3,769 samples. Models are commonly evaluated in terms of the mean Average Precision~(mAP) metric.
%with the 0.7 IoU threshold for the car class. 
mAP is calculated with recall 40 positions~(R40). 
We perform ablation studies with AP$_{\textrm{3D}}$~(R40) on the {\em val} split.
We conduct main comparisons with AP$_{\textrm{3D}}$~(R40) on {\em test} split and AP$_{\textrm{3D}}$~(R11) on the {\em val} split. For the optional multi-modal settings, RGB features are extracted from single front-view for fusion.

\vspace{0.5em}
\noindent
\textbf{nuScenes.}
The nuScenes~\cite{nuscenes} is a large-scale dataset, which contains 1,000 driving sequences in total. It is split into 700 scenes for training, 150 scenes for validation, and 150 scenes for testing. %nuScenes provides a rich data modality. 
It is collected using a 32-beam synced LIDAR and 6 cameras with the complete 360$^{\mathrm{o}}$ environment coverage. 
%Following the settings in our baseline method, CenterPoint~\cite{centerpoint}, we set the detection range to [-54m,~54m] for $X$ and $Y$ axes, and [-5m,~3m] for $Z$ axis, where the voxel size is set as (0.075m,~0.075m,~0.2m) for the corresponding axes. 
In evaluation, the main metrics are mAP and nuScenes detection score~(NDS). 
%We report the metrics mAP, NDS, and AP for all 10 classes. 
In terms of multi-modal experiments, we use images of 6 views for fusion. For ablation study, models are trained on $\frac{1}{4}$ training data and evaluated on the entire validation set, {\em i.e.}, nuScenes $\frac{1}{4}$ split.

\vspace{0.5em}
\noindent
\textbf{Implementation details.}
In experiments, we validate our modules on state-of-the-art frameworks of PV-RCNN~\cite{pvrcnn}, Voxel R-CNN~\cite{voxel-rcnn} on KITTI~\cite{kitti}, and CenterPoint~\cite{centerpoint} on nuScenes~\cite{nuscenes}. In LIDAR-only experiments, we apply \Ours in the first three stages of backbone networks. In multi-modal cases, we apply \OursF only in the first stage of the backbone network, for affordable memory and inference cost. We set the importance threshold $\tau$ to 0.5.
We keep other settings intact. More experimental details are provided in the {supplementary material}.
%({\em e.g.}, learning rate, optimizer, and others hyperparamters). %These details can be found in the {\em supplementary material}.

\begin{table}[t]
\begin{center}
\caption{Improvements on PV-RCNN in AP$_{\textrm{3D}}$(R40) on KITTI {\em val}.}
% \vspace{-0.5em}
\begin{threeparttable}
\resizebox{\linewidth}{!}{
\begin{tabular}{|l|cc|ccc|}
\hline
%        & \multirow{2}{*}{\makecell[c]{\#{\em Params}\\(M)}} & \multirow{2}{*}{\makecell[c]{{\em Time}\\(ms)}} & \multicolumn{3}{c|}{\em Car} \\
% $_{\textcolor[RGB]{34,139,34}{(+0.83)}}$
% $_{\textcolor[RGB]{34,139,34}{(+0.96)}}$
  {\em Method}       &  \#{\em Params}  &  {\em Runtime}   & Easy    & Mod.   & Hard    \\ \hline
PV-RCNN~\cite{pvrcnn}   & --    &  --  & 92.57   & 84.83      & 82.69   \\
PV-RCNN~$^{\circ}$ & 13.16M  & 103ms & 92.10 & 84.36 & 82.48 \\ \hline
\Ours  & 13.44M &  112ms  &  92.32  & 85.19      &  82.62  \\ 
\OursF  & 13.70M &  125ms  & 92.26   & \textbf{85.32}      & 82.95   \\ \hline
\end{tabular}
}
\begin{tablenotes}
  \small
  \item $\;\;^{\circ}$ These results are evaluated on the official released model.
\end{tablenotes}
\end{threeparttable}
\label{tab:improvements-pvrcnn}
\end{center}
\end{table}

\subsection{Ablation Studies}
\noindent
\textbf{Improvements on KITTI.}
We first evaluate our methods over PV-RCNN~\cite{pvrcnn} in Tab.~\ref{tab:improvements-pvrcnn}, as it is a high-performance, multi-class, and open-sourced framework. In Tab.~\ref{tab:improvements-pvrcnn}, 
the 1st and 2nd lines show the reported results~\cite{pvrcnn} and results tested from the released model. We take the latter as the baseline. Focal~S-Conv and \OursF achieve non-trivial improvement over this strong baseline. %Our multi-modal approach further improves this strong baseline.

\begin{table}[t]
\begin{center}
\caption{Ablations on dynamic shape in AP$_{\textrm{3D}}$ (R40) on KITTI {\em val}.}
% \vspace{-0.5em}
\resizebox{\linewidth}{!}{
\begin{tabular}{|l|c|ccc|c|c|}
\hline
        \multirow{2}{*}{\em Method}                    & \multicolumn{1}{c|}{\multirow{2}{*}{\em \makecell[c]{Dynamic\\shape}}} & \multicolumn{3}{c|}{\em Car} & {\em Ped.}      & {\em Cyc.}      \\
                            & \multicolumn{1}{c|}{}                        & Easy  & Mod. & Hard  & Mod. & Mod. \\ \hline
Baseline                     & \multicolumn{1}{c|}{--}                       & 92.10 & 84.36    & 82.48 & 54.49    & 70.38    \\ \hline
%\multirow{3}{*}{\Ours} & subm                                         & 91.61 & 82.84    & 82.13 & 60.44    & 72.71    \\
\multirow{2}{*}{\Ours} & \xmark                                      & 91.10 & 84.02    & 82.22 & 57.62    & 69.82    \\
                            & \cmark                                      & 92.32 & \textbf{85.19}    & 82.62 & \textbf{61.61}    & \textbf{72.76}    \\ \hline
\end{tabular}
}
\label{tab:ablation-dynamicoutput}
\end{center}
\end{table}
\begin{table}[t]
\begin{center}
\caption{Ablations on input selection in AP$_{\textrm{3D}}$ (R40) on KITTI {\em val}.}
% \vspace{-0.5em}
\resizebox{\linewidth}{!}{
\begin{tabular}{|l|c|ccc|c|c|}
\hline
          \multirow{2}{*}{\em Method}                  & \multicolumn{1}{c|}{\multirow{2}{*}{{\em \makecell[c]{Important\\selection}}}}         & \multicolumn{3}{c|}{\em Car} & {\em Ped.}      & {\em Cyc.}      \\
                            &                                 & Easy  & Mod. & Hard  & Mod. & Mod. \\ \hline
Baseline                     & \multicolumn{1}{c|}{--}                               & 92.10 & 84.36    & 82.48 & 54.49    & 70.38    \\ \hline
\multirow{2}{*}{\Ours} & \xmark     & 91.36 & 82.77    & 82.12 & 57.86    & 71.77    \\
                            & \cmark & 92.32 & \textbf{85.19}    & 82.62 & \textbf{61.61}    & \textbf{72.76}    \\ \hline
\end{tabular}
}
\label{tab:ablation-importancesampling}
\end{center}
\end{table}
\begin{table}[t]
\begin{center}
\caption{Ablations on supervisions in AP$_{\textrm{3D}}$ (R40) on KITTI {\em val}.}
% \vspace{-0.5em}
\resizebox{\linewidth}{!}{
\begin{tabular}{|l|l|ccc|c|c|}
\hline
          \multirow{2}{*}{\em Method}                  & \multicolumn{1}{c|}{\multirow{2}{*}{{\em Supervision}}}           & \multicolumn{3}{c|}{\em Car} & {\em Ped.}      & {\em Cyc.}      \\
                            &                                 & Easy  & Mod. & Hard  & Mod. & Mod. \\ \hline
Baseline                     & \multicolumn{1}{c|}{--}                               & 92.10 & 84.36    & 82.48 & 54.49    & 70.38    \\ \hline
\multirow{3}{*}{\Ours} & Attention           & 91.81 & 84.49    & 82.31 & 60.64    & \textbf{72.93}    \\
                            & Obj. loss & 92.39 & 85.05    & 82.62 &  59.27   & 71.46    \\ 
                    & Both & 92.32 & \textbf{85.19}    & 82.62 & \textbf{61.61}    & 72.76\\ \hline
\end{tabular}
}
\label{tab:ablation-objectiveloss}
\end{center}
\end{table}
\begin{table}[t]
\begin{center}
\caption{Ablations on use stages in AP$_{\textrm{3D}}$ (R40) on KITTI {\em val}.}
% \vspace{-0.5em}
\resizebox{\linewidth}{!}{
\begin{tabular}{|l|l|ccc|c|c|}
\hline
           \multirow{2}{*}{\em Method}                 & \multicolumn{1}{c|}{\multirow{2}{*}{\em Stages}} & \multicolumn{3}{c|}{\em Car} & {\em Ped.}      & {\em Cyc.}      \\
                            & \multicolumn{1}{c|}{}                        & Easy  & Mod. & Hard  & Mod. & Mod. \\ \hline
Baseline                     & \multicolumn{1}{c|}{--}                       & 92.10 & 84.36    & 82.48 & 54.49    & 70.38    \\ \hline
\multirow{4}{*}{\Ours} & (1,)                                         & 92.19 & 84.83    & 82.43 & 60.56    & 72.29    \\
                            & (1, 2)                                        & 91.95 & 84.95    & 82.67 & 60.17    & 72.74    \\
                            & (1, 2, 3)                                      & 92.32 & \textbf{85.19}    & 82.62 & \textbf{61.61}    & \textbf{72.76}    \\
                            & (1, 2, 3, 4)                                    & 91.96 & 84.42    & 82.31 & 60.33    & 72.53    \\ \hline
\end{tabular}
}
\label{tab:ablation-usingstages}
\end{center}
\end{table}
\begin{table}[t]
\begin{center}
\caption{Ablations on the importance threshold $\tau$ on KITTI {\em val}.}
% \vspace{-0.5em}
\resizebox{\linewidth}{!}{
\begin{tabular}{|c|c|c|c|c|c|}
\hline
Importance Threshold $\tau$ & 0.1   & 0.3   & 0.5   & 0.7   & 0.9   \\ \hline
AP$_{\textrm{3D}}$ (R40) - {\em Car}                  & 84.97 & 85.09 & 85.19 & 84.96 & 84.68 \\ \hline
\end{tabular}
}
\label{tab:importance-threshold}
\end{center}
\end{table}

%\vspace{0.5em}
\noindent
\textbf{Dynamic output shape.}
In \Ours, the output shape from every single voxel is dynamically determined by the predicted importance maps. We ablate this by fixing output shapes as regular dilation, without any other change. Tab.~\ref{tab:ablation-dynamicoutput} shows that dilating all sparse features is harmful. It dramatically increases the number of unpromising voxel features.

\vspace{0.5em}
\noindent
\textbf{Importance sampling.}
% gains importance sampling, replace it with random sampling (KITTI) [Done]
\Ours selects sparse features that need dilation with predicted importance. To ablate this module, we replace the importance selection (the {important input selection} step) with a random sample in Tab.~\ref{tab:ablation-importancesampling} without other changes. It shows that large performance drop occurs without the guidance of importance. This validates that the importance prediction is necessary.

\begin{table}[t]
\begin{center}
\caption{Comparison on KITTI {\em test} split in AP$_{\textrm{3D}}$ (R40) for {\em Car}.}
% \vspace{-0.5em}
\resizebox{\linewidth}{!}{
\begin{tabular}{|l|c|ccc|}
\hline
%\multicolumn{1}{|c|}{\multirow{2}{*}{\em Method}} & \multirow{2}{*}{\em Fusion} & \multicolumn{3}{c|}{AP$_{\textrm{3D}}$(R40)} \\
{\em Method}                        &       {\em Fusion}                       & Easy    & \textbf{Mod.}    & Hard   \\ \hline \hline
MV3D~\cite{mutliview3ddet}                                          & \multirow{8}{*}{\cmark}            & 74.97   & 63.63  & 54.00  \\
%ContFuse~\cite{deepfusion}                                      &                              & 83.68   & 68.78  & 61.67  \\
F-PointNet~\cite{fpointnet}                                    &                              & 82.19   & 69.79  & 60.59  \\
AVOD-FPN~\cite{joint3dproposal}                                      &                              & 83.07   & 71.76  & 65.73  \\
PointSIFT+SENet~\cite{pointsiftsenet}                               &                              & 85.99   & 72.72  & 64.58  \\
MMF~\cite{multitaskfusion}                                           &                              & 88.40   & 77.43  & 70.22  \\
EPNet~\cite{epnet}                                           &                              & 89.81   & 79.28  & 74.59  \\
3D-CVF~\cite{3dcvf}                                        &                              & 89.20   & 80.05  & 73.11  \\
CLOCs~\cite{clocs}                                         &                              & 88.94   & 80.67  & 77.15  \\ \hline \hline
%VoxelNet~\cite{voxelnet}                                      & \multirow{14}{*}{\xmark}           & 77.47   & 65.11  & 57.73  \\
%SECOND~\cite{second}                                        &                              & 83.34   & 72.55  & 65.82  \\
PointPillars~\cite{pointpillars}                                  &          \multirow{10}{*}{\xmark}                    & 82.58   & 74.31  & 68.99  \\
Point R-CNN~\cite{point-rcnn}                                   &                              & 86.96   & 75.64  & 70.70  \\
%TANet~\cite{tanet}                                         &                              & 85.94   & 75.76  & 68.32  \\
%HVNet~\cite{hvnet}                                         &                              & 87.21   & 77.58  & 71.79  \\
Part-$A^2$~\cite{part-a2}                                     &                              & 87.81   & 78.49  & 73.51  \\
%3DSSD~\cite{3dssd}                                         &                              & 88.36   & 79.57  & 74.55  \\
STD~\cite{std}                                           &                              & 87.95   & 79.71  & 75.09  \\
SA-SSD~\cite{sassd}                                        &                              & 88.75   & 79.79  & 74.16  \\
PV-RCNN~\cite{pvrcnn}                                       &                              & 90.25   & 81.43  & 76.82  \\
Pyramid-PV~\cite{pyramid-rcnn}                                    &                              & 88.39   & 82.08  & 77.49  \\
VoTr-TSD~\cite{voxeltransformer}                                      &                              & 89.90   & 82.09  & 79.14  \\ \hline \hline
Voxel R-CNN~\cite{voxel-rcnn}                                   &     \xmark                          & 90.90   & 81.62  & 77.06  \\
\Ours                                    &     \xmark                    & 90.20   & 82.12  & 77.50  \\
% \textcolor[RGB]{34,139,34}{$_{(+0.50)}$}, \textcolor[RGB]{34,139,34}{$_{(+0.66)}$}
\OursF                        &     \cmark                         & 90.55   & \textbf{82.28}  & 77.59  \\ \hline
\end{tabular}
}
\label{tab:kitti-test}
\end{center}
\end{table}
\vspace{0.5em}
\noindent
\textbf{Supervision setting.}
% gains from loss function (KITTI) [Done]
The additional branch in \Ours is supervised by both attention multiplication and the objective loss. We ablate them in Tab.~\ref{tab:ablation-objectiveloss}. Only using objective loss supervision is enough to ensure performance on {\em Car}. However, its performance on minor classes, {\em Ped.} and {\em Cyc.}, is not optimal. Attention multiplication is beneficial to {\em Ped.} and {\em Cyc}. We assume that minor classes cannot get balanced supervision from the objective loss like the long-tailed distribution. In contrast, attention multiplication is object-agnostic, relaxing the imbalance to some degree.

\vspace{0.5em}
\noindent
\textbf{Stages for using focal sparse convolution.}
% number of focal sparse convolution to use (KITTI) [Done]
Tab.~\ref{tab:ablation-usingstages} shows results of using \Ours in different numbers of stages. (1) Applying \Ours in the first stage, which already obtains clear improvement. The performance enhances as the used stage increases until all stages are involved. Since \Ours adjusts output sparsity, it is reasonable to be used in early stages that make effects on subsequent feature learning. The spatial feature space in the last stage is down-sampled to a very limited size, which might not be large enough for sparsity adaptation. Empirically, usage in the last layer of the first three stages is the best choice. It is thus used as the default setting in our experiments. 

\begin{table}[t]
\begin{center}
\caption{Comparison on KITTI {\em val} split in AP$_{\textrm{3D}}$ (R11) for {\em Car}.}
% \vspace{-0.5em}
\resizebox{\linewidth}{!}{
\begin{tabular}{|l|c|ccc|}
\hline
%\multicolumn{1}{|c|}{\multirow{2}{*}{\em Method}} &\multicolumn{1}{|c|}{\multirow{2}{*}{\em Fusion}} & \multicolumn{3}{c|}{AP$_{\textrm{3D}}$ ({R11})}                     \\
{\em Method}             &   {\em Fusion}        & Easy          & \textbf{Mod.}           & Hard          \\ \hline \hline
F-PointNet~\cite{fpointnet}             &     \multirow{3}{*}{\cmark}                & 83.76         & 70.92         & 63.65         \\ 
%AVOD-FPN~\cite{joint3dproposal}             &                     & -         & 74.4         & -         \\
PointSIFT+SENet~\cite{pointsiftsenet}  &          &     85.62     &      72.05    &    64.19      \\ 
3D-CVF~\cite{3dcvf}  &          &     89.67     &      79.88    &    78.47      \\ 
\hline \hline
%VoxelNet~\cite{votenet}                                      & 81.97         & 65.46         & 62.85         \\
PointPillars~\cite{pointpillars}             &     \multirow{9}{*}{\xmark}                & 86.62         & 76.06         & 68.91         \\
%TANet~\cite{tanet}                                         & 87.52         & 76.64         & 73.86         \\
%SECOND~\cite{second}                                        & 88.61         & 78.62         & 77.22         \\
Point R-CNN~\cite{point-rcnn}                &                   & 88.88         & 78.63         & 77.38         \\
Part-$A^2$~\cite{part-a2}                     &                & 89.47         & 79.47         & 78.54         \\
%3DSSD~\cite{3dssd}                                         & 89.71         & 79.45         & 78.67         \\
STD~\cite{std}                               &            & 89.70         & 79.80         & 79.30         \\
SA-SSD~\cite{sassd}                          &           & 90.15         & 79.91         & 78.78         \\
Deform. PV-RCNN~\cite{deformable-pvrcnn}                                       &                              &  -  & 83.30  & -  \\
PV-RCNN~\cite{pvrcnn}                        &               & 89.35         & 83.69         & 78.70         \\
VoTr-TSD~\cite{voxeltransformer}             &                         & 89.04         & 84.04         & 78.68         \\
Pyramid-PV~\cite{pyramid-rcnn}               &                     & 89.37         & 84.38         & 78.84         \\ \hline \hline
Voxel R-CNN~\cite{voxel-rcnn}                &    \xmark               & 89.41         & 84.52         & 78.93         \\
% \textcolor[RGB]{34,139,34}{$_{(+0.11)}$}, \textcolor[RGB]{34,139,34}{$_{(+0.41)}$}, \textcolor[RGB]{34,139,34}{$_{(+0.25)}$}
% \textcolor[RGB]{34,139,34}{$_{(+0.41)}$}, \textcolor[RGB]{34,139,34}{$_{(+0.70)}$}, \textcolor[RGB]{34,139,34}{$_{(+6.26)}$}
\Ours                   &      \xmark           & 89.52  & 84.93  & 79.18  \\
\OursF                  &      \cmark        & 89.82  & \textbf{85.22}  & 85.19  \\ \hline
\end{tabular}
}
\label{tab:kitti-val}
\end{center}
\end{table}

\begin{table}[t]
\begin{center}
\caption{Improvement over multi-modal baseline on nuScenes $\frac{1}{4}$.}
% \vspace{-0.5em}
\resizebox{\linewidth}{!}{
\begin{tabular}{|l|cc|cc|}
\hline

                              & \#{\em Params} & {\em Runtime} & mAP   & NDS                  \\ \hline
CenterPoint      & 9.0M & 93ms & 56.1  & 64.2    \\ 
+ Fusion   & 9.24M & 145ms &  59.0 \textcolor[RGB]{34,139,34}{$_{(+2.9)}$}  & 65.6 \textcolor[RGB]{34,139,34}{$_{(+1.4)}$}       \\ 
\OursF   & 9.25M & 159ms & \textbf{61.7} \textcolor[RGB]{34,139,34}{$_{(+5.6)}$}                      & \textbf{67.2} \textcolor[RGB]{34,139,34}{$_{(+2.9)}$}    \\  \hline
\end{tabular}}
\label{tab:improvements-multimodal-1/4}
\end{center}
\end{table}
\begin{table}[t]
\begin{center}
\caption{Ablations on use stage and fusion scope on nuScenes $\frac{1}{4}$.}
% \vspace{-0.5em}
\resizebox{\linewidth}{!}{
\begin{tabular}{|c|c|ccc|ccc|}
\hline
$\;${\em Stage}$\;$  & None & \multicolumn{3}{c|}{1} & 2    & 3    & 4    \\
$\;${\em Scope}$\;$ & -    & None   & Imp.  & All   & Imp. & Imp. & Imp. \\ \hline
mAP          & 56.1 & 58.6   & \textbf{61.7}  & 60.9  & 60.7 & 55.0 & 54.8 \\
NDS          & 64.2 & 66.2   & \textbf{67.2}  & 66.4  & 66.5 & 63.5 & 63.3 \\ \hline
\end{tabular}
}
\label{tab:improvements-multimodal-stage}
\end{center}
\end{table}

\begin{table*}[t]
\begin{center}
\caption{Comparison with other methods on nuScenes {\em test} split.}
% \vspace{-0.5em}
\begin{threeparttable}
\resizebox{\linewidth}{!}{
\begin{tabular}{|l|c|cc|cccccccccc|}
\hline
                   {\em Method}                 & {\em Fusion} & mAP   & NDS & Car                  & Truck                & Bus                  & Trailer              & C.V.                   & Ped                  & Mot                  & Byc                  & T.C.                   & Bar                   \\ \hline \hline
PointPillars~\cite{pointpillars}                        & \xmark & 30.5 & 45.3 & 68.4 & 23.0 & 28.2 & 23.4 & 4.1 & 59.7 & 27.4 & 1.1 & 30.8 & 38.9 \\ 
3DSSD~\cite{3dssd}                         & \xmark & 42.6 & 56.4 &  81.2 & 47.2 & 61.4 & 30.5 & 12.6 & 70.2 & 36.0 & 8.6 & 31.1 & 47.9 \\ 
CBGS~\cite{cbgs}                        & \xmark & 52.8 & 63.3 & 81.1 & 48.5 & 54.9 & 42.9 & 10.5 & 80.1 & 51.5 & 22.3 & 70.9 & 65.7 \\ 
HotSpotNet~\cite{hotspotnet}                   & \xmark & 59.3 & 66.0 & 83.1 & 50.9 & 56.4 & 53.3 & 23.0 & 81.3 & 63.5 & 36.6 & 73.0 & 71.6 \\
CVCNET~\cite{cvcnet}                  & \xmark & 58.2 & 66.6  & 82.6 & 49.5 & 59.4 & 51.1 & 16.2 & 83.0 & 61.8 & 38.8 & 69.7 & 69.7 \\ 
\hline \hline
PointPainting~\cite{pointpainting}                         & \cmark & 46.4 & 58.1 & 77.9 & 35.8 & 36.2 & 37.3 & 15.8 & 73.3 & 41.5 & 24.1 & 62.4 & 60.2 \\ 
3DCVF~\cite{3dcvf}                        & \cmark & 52.7 & 62.3 & 83.0 & 45.0 & 48.8 & 49.6 & 15.9 & 74.2 & 51.2 & 30.4 & 62.9 & 65.9 \\ 
FusionPainting~\cite{fusionpainting} & \cmark & 66.3 & 70.4 & 86.3 & 58.5 & 66.8 & 59.4 & 27.7 & 87.5 & 71.2 & 51.7 & 84.2 & 70.2 \\ 
MVF~\cite{mvf}                         & \cmark & 66.4 & 70.5 & 86.8 & 58.5 & 67.4 & 57.3 & 26.1 & 89.1 & 70.0 & 49.3 & 85.0 & 74.8 \\
PointAugmenting~\cite{pointaugmenting}                        & \cmark & 66.8 & 71.0 & 87.5 & 57.3 & 65.2 & 60.7 & 28.0 & 87.9 & 74.3 & 50.9 & 83.6 & 72.6 \\
\hline \hline
CenterPoint~\cite{centerpoint}                         & \xmark & 58.0 & 65.5 & 84.6 & 51.0 & 60.2 & 53.2 & 17.5 & 83.4 & 53.7 & 28.7 & 76.7 & 70.9 \\ 
CenterPoint$^{\dagger}$                         & \xmark & 60.3 & 67.3 & 85.2 & 53.5 & 63.6 & 56.0 & 20.0 & 84.6 & 59.5 & 30.7 & 78.4 & 71.1 \\ 
CenterPoint v2$^{\star}$                         & \cmark & 67.1 & 71.4 & 87.0 & 57.3 & 69.3 & 60.4 & 28.8 & 90.4 & 71.3 & 49.0 & 86.8 & 71.0 \\\hline
\Ours                        & \xmark & 63.8 & 70.0 & 86.7 & 56.3 & 67.7 & 59.5 & 23.8 & 87.5 & 64.5 & 36.3 & 81.4 & 74.1 \\ 
\OursF                        & \cmark & 67.8 & 71.8 & 86.5 & 57.5 & 68.7 & 60.6 & 31.2 & 87.3 & 76.4 & 52.5 & 84.6 & 72.3 \\ 
\OursF$^{\dagger}$                        & \cmark & 68.9 & 72.8 & 86.9 & 59.3 & 68.7 & 62.5 & 32.8 & 87.8 & 78.5 & 53.9 & 85.5 & 72.8 \\  
\OursF$^{\ddagger}$                        & \cmark & \textbf{70.1} & \textbf{73.6} & 87.5 & 60.0 & 69.9 & 64.0 & 32.6 & 89.0 & 81.1 & 59.2 & 85.5 & 71.8 \\  \hline
\end{tabular}}
\begin{tablenotes}
  \small
  \item $\;\;^{\dagger}$ Flip testing.
  $\;\;^{\ddagger}$ Flip and rotation testing.
  $^{\;\;\star}$ CenterPoint v2 includes PointPainting with Cascade R-CNN~\cite{cascade-rcnn} and model-ensembling.
\end{tablenotes}
\end{threeparttable}
\label{tab:nuscenes-test}
\end{center}
\end{table*}

\vspace{0.5em}
\noindent
\textbf{Importance threshold.} We ablate the importance threshold $\tau$ used in \Ours in Tab.~\ref{tab:importance-threshold}. We run experiments with this value ranging from 0.1 to 0.9 and interval 0.2, without other change of settings. The accuracy AP$_{3D}$ (R40) on {\em Car} serves as the metric in this ablation.
The performance is stable as the threshold value $\tau$ varies. 

\vspace{0.5em}
\noindent
\textbf{Improvements over multi-modal baseline on nuScenes.}
% further gains from multi-modal focal sparse convolution (KITTI + 1/4 nuScenes)
We evaluate our multi-modal \Ours on the nuScenes~\cite{nuscenes} 1/4 dataset. More improvement is presented in Tab.~\ref{tab:improvements-multimodal-1/4}. We build a multi-modal CenterPoint baseline by fusing image features to the same fusion layer used in our methods, with the same fusion and feature extraction layers. This multi-modal CenterPoint enhances the LIDAR-only baseline from 56.1\% to 59.0\% mAP. \OursF improves to 61.7\% mAP on this strong baseline.

\vspace{0.5em}
\noindent
\textbf{Use stages and fusion scope for \OursF.}
% positions to use multi-modal focal sparse convolution (1/4 nuScenes) [Done]
We ablate the usage stages and fusion scope for \OursF in Tab.~\ref{tab:improvements-multimodal-stage}. Fusion scope is the scope of sparse features to fuse with RGB features at the output of \OursF. It shows that fusion in the early stages is beneficial, and becomes adverse in the last two stages. {\em Imp.} means only fusing onto important output features (judged by importance maps). When fusing in the first stage, it is better to fuse on {\em important} features, instead of all of them, making representation discriminative.

% visualization on both images and bev (nuScenes) [TODO]

\vspace{0.5em}
\noindent
\textbf{Model complexity and runtime.}
We report the model complexity and runtime comparisons in Tab.~\ref{tab:improvements-pvrcnn} and \ref{tab:improvements-multimodal-1/4}. The runtimes are evaluated on the same GPU machine. \Ours and its multi-modal variant only add a small overhead to model parameters and computation, on KITTI~\cite{kitti}. This indicates that the performance improvement comes from the model capacity of sparsity learning, instead of increasing model sizes. On nuScenes~\cite{nuscenes}, the overall runtime rises from 93 ms to 159 ms. But parameters are still limited. It is a common limitation in multi-view fusion methods. The multi-modal baseline also requires 145 ms. The reason is that there are 6-view images to process per frame. 

\subsection{Main Results}
% KITTI and nuScenes, LIDAR-only and multi-modal

%\vspace{0.5em}
\noindent
\textbf{KITTI.}
% KITTI test AP R40, val AP R11
We compare our \Ours modules upon Voxel R-CNN~\cite{voxel-rcnn} with previous state-of-the-art methods on both the KITTI {\em test} and {\em val} split. In Tab.~\ref{tab:kitti-test}, we compare with both LIDAR-only and multi-modal methods. The original Voxel R-CNN~\cite{voxel-rcnn} is comparable to PV-RCNN~\cite{pvrcnn} and is inferior to Pyramid-PV~\cite{pyramid-rcnn} and VoTr-TSD~\cite{voxeltransformer}. \Ours improves it to surpass these two new methods. Using \OursF, the multi-modal Voxel R-CNN achieves 82.28\% AP$_{3D}$ on the KITTI {\em test} split. Tab.~\ref{tab:kitti-val} shows comparisons on KITTI {\em val} split in AP$_{3D}$ in recall 11 positions. 
%Voxel R-CNN~\cite{voxel-rcnn} already achieves state-of-the-art performance.
\Ours and \OursF enhance this leading result to 84.93\% and 85.22\% respectively in {\em Car} class. 

\vspace{0.5em}
\noindent
\textbf{nuScenes.}
% nuScenes test [TODO]
On the nuScenes dataset, we evaluate our models on the test server and compare them with both LIDAR-only and multi-modal methods, as in Tab.~\ref{tab:nuscenes-test}. \Ours improves CenterPoint~\cite{centerpoint} by a large margin to 63.8\% mAP. Multi-modal methods present much better performance than LIDAR-only methods on the nuScenes dataset. 
%PointAugmenting~\cite{pointaugmenting}, also based on CenterPoint~\cite{centerpoint}, achieves previous state-of-the-art performance, while it requires a 2D detection model to enrich raw LIDAR data.
CenterPoint v2$^{\star}$ includes PointPainting~\cite{pointpainting}, Cascade R-CNN~\cite{cascade-rcnn} instance segmentation models pre-trained on nuImages, and five-model ensembling.
As the testing augmentations are not unified or stated in previous methods, we provide two results of our final model.
\OursF achieves 67.8\% mAP and 71.8\% mAP without any ensembling or testing augmentation. \OursF$^{\ddagger}$ further achieves 70.1\% mAP and 73.6\% NDS with test-time augmentations~\cite{centerpoint}. Both results outperform previous methods.

%------------------------------------------------------------------------
\section{Conclusion and Discussion}
\label{sec:conclusion}
This paper presents a focal sparse convolution and a multi-modal extension, which are simple and effective. They are end-to-end solutions for LIDAR-only and multi-modal 3D object detection. For the first time, we show that the learned sparsity with focal points is essential for 3D object detectors. Notably, focal and fusion sparse CNNs achieve leading performance on the large-scale nuScenes.

\vspace{0.35em}
\noindent
\textbf{Limitations.}
In the multi-modal 3D detection that requires multiple views, {\em e.g.}, 6 high-resolution images per frame in nuScenes~\cite{nuscenes}, computation cost increases, although the image branch is already largely simplified.

\noindent
\textbf{Boarder Impacts.}
The proposed method replies on the sparsity of data distribution. It might reflect biases in data collection, including the ones of negative societal impacts.

%\vspace{0.35em}
\noindent
\textbf{Acknowledgements.} This work is in part supported by The National Key Research and Development Program of China (No. 2017YFA0700800) and Beijing Academy of Artificial Intelligence (BAAI).

%%%%%%%%% REFERENCES
{\small
\bibliographystyle{ieee_fullname}
\bibliography{egbib}
}

\appendix
\captionsetup[table]{labelformat={default},labelsep=period,name={Table S -}}
\captionsetup[figure]{labelformat={default},labelsep=period,name={Figure S -}}

%\title{Focal Sparse Convolutional Networks for 3D Object Detection \\
%{\em Supplementary Material}}
\section*{\Large Appendix}

%%%%%%%%% BODY TEXT
\section{More Implementation Details}
\label{sec:training-details}
Our implementation is based on the open-sourced OpenPCDet~\cite{pvrcnn, voxel-rcnn}, and the released code of CenterPoint~\cite{centerpoint}.

\subsection{Voxelization}
\noindent
\textbf{KITTI.}
The 3D object detectors in this work convert point clouds into voxels as input data. 
On the KITTI~\cite{kitti} dataset, the range of point clouds is clipped into [0, 70.4m] for {\em X} axis, [-40m,40m] for {\em Y} axis, and [-3, 1]m for {\em Z} axis. The voxelization size for input is (0.05m, 0.05m, 0.1m). 

\vspace{0.5em}
\noindent
\textbf{nuScenes.}
On the nuScenes~\cite{nuscenes}, the detection range is set to [-54m, 54m] for both {\em X} and {\em Y} axes, and [-5m, 3m] for the {\em Z} axis. The voxel size is set as (0.075m, 0.075m, 0.2m).

\subsection{Data Augmentations}
\noindent
\textbf{KITTI.}
On the KITTI~\cite{kitti} dataset, data transformation and augmentations include random flipping, global scaling, global rotation, and ground-truth (GT) sampling~\cite{second}. The random flipping is conducted along the {\em X} axis. The global scaling factor is sampled from 0.95 to 1.05. The global rotation is conducted around the {\em Z} axis. The rotation angle is sampled from -45$^{\mathrm{o}}$ and 45$^{\mathrm{o}}$. The ground-truth sampling is to copy-paste some new objects from other scenes to the current training data, which enriches objects in the environments. For the multi-modal setting, we do not transform images with the corresponding operations, except ground-truth sampling. We copy-paste the corresponding image crops from other scenes onto the current training images.

\vspace{0.5em}
\noindent
\textbf{nuScenes.}
On the nuScenes~\cite{nuscenes} dataset, data augmentations includes random flipping, global scaling, global rotation, GT sampling~\cite{second}, and an additional translation. The random flipping is conducted along both {\em X} and {\em Y} axes. The rotation angle is also randomly sampled in [-45$^{\mathrm{o}}$, 45$^{\mathrm{o}}$]. The global scaling factor is sampled in [0.9, 1.1]. The translation noise is conducted on all three axes, {\em X}, {\em Y}, and {\em Z}, with a factor independently sampled from 0 to 0.5. We also conduct the corresponding point-image GT sampling on the nuScenes. GT sampling is disabled in the last 4 epochs~\cite{pointaugmenting} for performance enhancement.

\begin{table}[t]
\begin{center}
\caption{Comparisons to Voxel R-CNN in R40 on KITTI {\em val}.}
\resizebox{\linewidth}{!}{
\begin{tabular}{|l|ccc|ccc|}
\hline
          \multirow{2}{*}{\em Method}                  & \multicolumn{3}{c|}{AP$_{\textrm{BEV}}$} & \multicolumn{3}{c|}{AP$_{\textrm{3D}}$}    \\
                            & Easy  & Mod. & Hard  & Easy & Mod. & Hard \\ \hline
V.~\cite{voxel-rcnn} & 95.52 & 91.25 & 88.99 & 92.38 & 85.29 & 82.86 \\
Ours        & 95.45 & \textbf{91.51} & 91.21 & 92.86 & \textbf{85.85} & 85.29 \\ \hline
\end{tabular}
}
\label{tab:kitti-val-r40}
\end{center}
\end{table}
\begin{table}[t]
\begin{center}
\caption{Objective loss weight in AP$_{\textrm{3D}}$ (R40) on KITTI {\em val}.}
% \vspace{-0.5em}
\resizebox{\linewidth}{!}{
\begin{tabular}{|l|c|ccc|c|c|}
\hline
           \multirow{2}{*}{\em Method}                 & \multicolumn{1}{c|}{\multirow{2}{*}{\em Weight}} & \multicolumn{3}{c|}{\em Car} & {\em Ped.}      & {\em Cyc.}      \\
                            & \multicolumn{1}{c|}{}                        & Easy  & Mod. & Hard  & Mod. & Mod. \\ \hline
Baseline                     & \multicolumn{1}{c|}{--}                       & 92.10 & 84.36    & 82.48 & 54.49    & 70.38    \\ \hline
\multirow{4}{*}{\makecell[c]{Focals\\Conv}} & 0.1 & 91.59 & 84.63 & 82.42 & 60.62 & 71.33 \\
                            & 0.5 & 92.10 & 85.16 & 83.12 & \textbf{63.74} & 69.56   \\
                            & 1.0   & 92.32 & \textbf{85.19}    & 82.62 & 61.61    & \textbf{72.76}    \\
                            & 2.0  & 91.73 & 84.61 & 82.38 & 54.09 & 71.34 \\ \hline
\end{tabular}
}
\label{tab:ablation-lossweight}
\end{center}
\end{table}
\begin{table}[t]
\begin{center}
\caption{Improvements upon CenterPoint on Waymo $\frac{1}{5}$.}
\resizebox{\linewidth}{!}{
\begin{tabular}{|l|ccc|ccc|}
\hline
          \multirow{2}{*}{\em Method}                  & \multicolumn{3}{c|}{AP LEVEL 1} & \multicolumn{3}{c|}{AP LEVEL 2} \\
                            & Veh.  & Ped. & Cyc.  & Veh.  & Ped. & Cyc. \\ \hline
Baseline & 70.9 & 71.5 & 69.1 & 62.8 & 63.5 & 66.5 \\
Focals Conv        & 72.2 & 72.6 & 71.1 & 64.1 & 64.6 & 68.5 \\ \hline
\end{tabular}
}
\label{tab:waymo}
\end{center}
\end{table}
\subsection{Training Settings}
\noindent
\textbf{KITTI.}
For model training on the KITTI dataset, {\em i.e.}, PV-RCNN~\cite{pvrcnn} and Voxel R-CNN~\cite{voxel-rcnn}, we train the network for 80 epochs with the batch size 16. We adopt the Adam optimizer. The learning rate is set as 0.01 and decreases in the cosine annealing strategy. The weight decay is set as 0.01. The momentum is set as 0.9. The gradient norms of training parameters are clipped by 10.

\vspace{0.5em}
\noindent
\textbf{nuScenes.}
For models trained on the nuScenes datasets, {\em i.e.}, CenterPoint~\cite{centerpoint}, we also train the network for 20 epochs with batch size 32. They are also trained with the Adam optimizer. The learning rate is initialized as 1e-3 and decreases in the cosine annealing strategy to 1e-4. The weight decay is set as 0.01. The gradient norms of training parameters are clipped by 35.

\begin{figure*}[t]
\begin{center}
   \includegraphics[width=\linewidth]{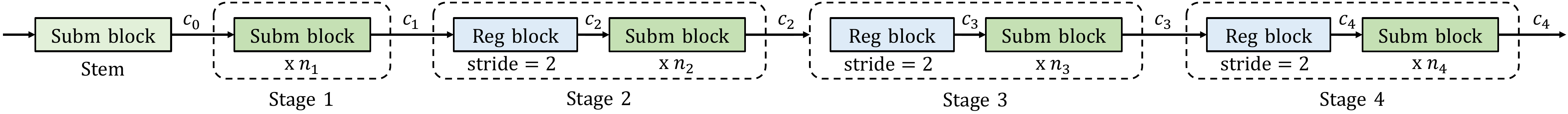}
   \caption{The illustration of the VoxelNet~\cite{voxelnet} backbone networks of PV-RCNN~\cite{pvrcnn}, Voxel R-CNN~\cite{voxel-rcnn}, and CenterPoint~\cite{centerpoint}. \{$c_0$, $c_1$, $c_2$, $c_3$, $c_4$\} represent the output channels of the Stem, Stage 1, 2, 3, and 4. \{$n_1$, $n_2$, $n_3$, $n_4$\} means the numbers of repeated submanifold blocks in these stages. In our approach, for the LIDAR-only setting, focal sparse convolutions (Focals Conv) are used in the last layer of Stage 1, 2, and 3. For the multi-modal setting, the focal sparse convolution with fusion (Focals Conv - F) is used in the last layer of Stage~1.}
   \label{fig:backbone-network}
\end{center}
\end{figure*}
\begin{table*}[t]
\begin{center}
\caption{Improvements over CenterPoint on the nuScenes {\em val} split.}
% \vspace{-0.5em}
\resizebox{0.9\linewidth}{!}{
\begin{tabular}{|l|cc|cccccccccc|}
\hline
                   {\em Method}                 & mAP   & NDS & Car                  & Truck                & Bus                  & Trailer              & C.V.                   & Ped                  & Mot                  & Byc                  & T.C.                   & Bar                   \\ \hline \hline
CenterPoint     & 59.0 & 66.4 & 85.6 & 57.2 & 71.2 & 37.3 & 16.2 & 85.1 & 58.4 & 41.0 & 69.2 & 68.2 \\ 
%CenterPoint~\cite{centerpoint}     & 61.1 & 68.3 & 86.5 & 60.4 & 72.5 & 40.4 & 19.8 & 86.0 & 61.1 & 45.2 & 70.8 & 68.9 \\
\Ours & 61.2 & 68.1 & 86.6 & 60.2 & 72.3 & 40.8 & 20.1 & 86.2 & 61.3 & 45.6 & 70.2 & 69.3 \\
%\Ours           & 62.3 & 69.0 & 86.5 & 61.2 & \textbf{73.5} & 40.3 & 21.3 & 86.5 & 63.3 & 48.0 & 72.0 & \textbf{70.6} \\
\OursF          & 63.9 & 69.4 & 86.5 & 58.5 & 72.4 & 41.2 & 23.9 & 86.0 & 69.0 & 55.2 & 76.9 & 69.1 \\

- GT-S. last 4  & 65.6 & 70.4 & 87.1 & 62.1 & \textbf{72.7} & 42.6 & 27.1 & 86.5 & 72.4 & 61.0 & 78.1 & 66.3 \\

+ Flip testing  & \textbf{67.1} & \textbf{71.5} & \textbf{87.7} & \textbf{62.9} & 72.4 & \textbf{42.6} & \textbf{28.1} & \textbf{87.8} & \textbf{74.4} & \textbf{65.5} & \textbf{78.9} & \textbf{70.1} \\  \hline
\end{tabular}}
\label{tab:improvements-over-centerpoint}
\end{center}
\end{table*}

\begin{table*}[t]
\begin{center}
\caption{Ablations on ground-truth sampling on nuScenes $\frac{1}{4}$. GT Sampl. - Ground-truth Sampling. Trans. - Transformations.}
\resizebox{0.9\linewidth}{!}{
\begin{tabular}{|c|cc|c|cccccccccc|}
\hline
Fusion & \makecell[c]{GT sampl.} & Trans. & mAP           & Car           & Truck         & Bus           & Trail.        & C.V.          & Ped           & Mot           & Byc           & T.C.          & Bar           \\ \hline
\xmark   & \multirow{2}{*}{\cmark}    & \multirow{2}{*}{\xmark}     & 39.3          & \textbf{70.8} & 31.0          & \textbf{49.0} & 20.3          & 3.5           & \textbf{73.7} & 28.8          & 13.4          & 49.3          & \textbf{50.6} \\
\cmark   &     &      & \textbf{43.3} & \textbf{70.8} & \textbf{32.5} & 47.9          & \textbf{21.0} & \textbf{6.7}  & 72.6          & \textbf{44.9} & \textbf{31.8} & \textbf{58.4} & 49.1          \\ \hline
\xmark   & \multirow{2}{*}{\xmark}    & \multirow{2}{*}{\cmark}     & 54.6          & 80.4          & 51.9          & 60.6          & 31.5          & 14.2          & 81.4          & 57.4          & 45.4          & 63.1          & 60.7          \\
\cmark   &     &      & \textbf{59.0} & \textbf{83.2} & \textbf{56.1} & \textbf{61.5} & \textbf{36.6} & \textbf{19.7} & \textbf{84.3} & \textbf{59.1} & \textbf{49.4} & \textbf{73.8} & \textbf{66.3} \\ \hline
\end{tabular}}
\label{tab:nuscenes-1/4-augs}
\end{center}
\end{table*}

\begin{table*}[t]
\begin{center}
\caption{Ablations on voxel size upon \OursF on the nuScenes {\em val} split.}
% \vspace{-0.5em}
\resizebox{0.9\linewidth}{!}{
\begin{tabular}{|l|cc|cccccccccc|}
\hline
                Voxel size (m) & mAP   & NDS & Car                  & Truck                & Bus                  & Trailer              & C.V.                   & Ped                  & Mot                  & Byc                  & T.C.                   & Bar                   \\ \hline \hline
(0.05, 0.05, 0.2) & 65.6 & 70.2 & 85.9 & 61.6 & 70.1 & 35.9 & 25.0 & \textbf{88.3} & 74.0 & \textbf{66.1} & \textbf{79.4} & \textbf{70.1} \\ 
                    (0.075, 0.075, 0.2) & \textbf{67.1} & \textbf{71.5} & \textbf{87.7} & \textbf{62.9} & 72.4 & 42.6 & 28.1 & 87.8 & 74.4 & 65.5 & 78.9 & \textbf{70.1} \\ 
            (0.1, 0.1, 0.2) & 66.5 & 71.4 & 87.5 & 62.1 & 71.8 & 44.6 & 27.9 & 86.9 & 74.3 & 63.6 & 77.4 & 69.0 \\ 
            (0.125, 0.125, 0.2) & 66.6 & 70.9 & 87.0 & 61.1 & \textbf{74.0} & 44.1 & \textbf{30.2} & 85.6 & \textbf{75.1} & 63.8 & 75.1 & 69.6 \\ 
            (0.15, 0.15, 0.2) & 65.3 & 70.2 & 86.9 & 60.9 & 72.8 & \textbf{45.1} & 30.0 & 83.3 & 70.4 & 63.5 & 72.7 & 67.9 \\
\hline
\end{tabular}}
\label{tab:ablations-voxelsize}
\end{center}
\end{table*}

\begin{figure}[t]
\begin{center}
   \includegraphics[width=\linewidth]{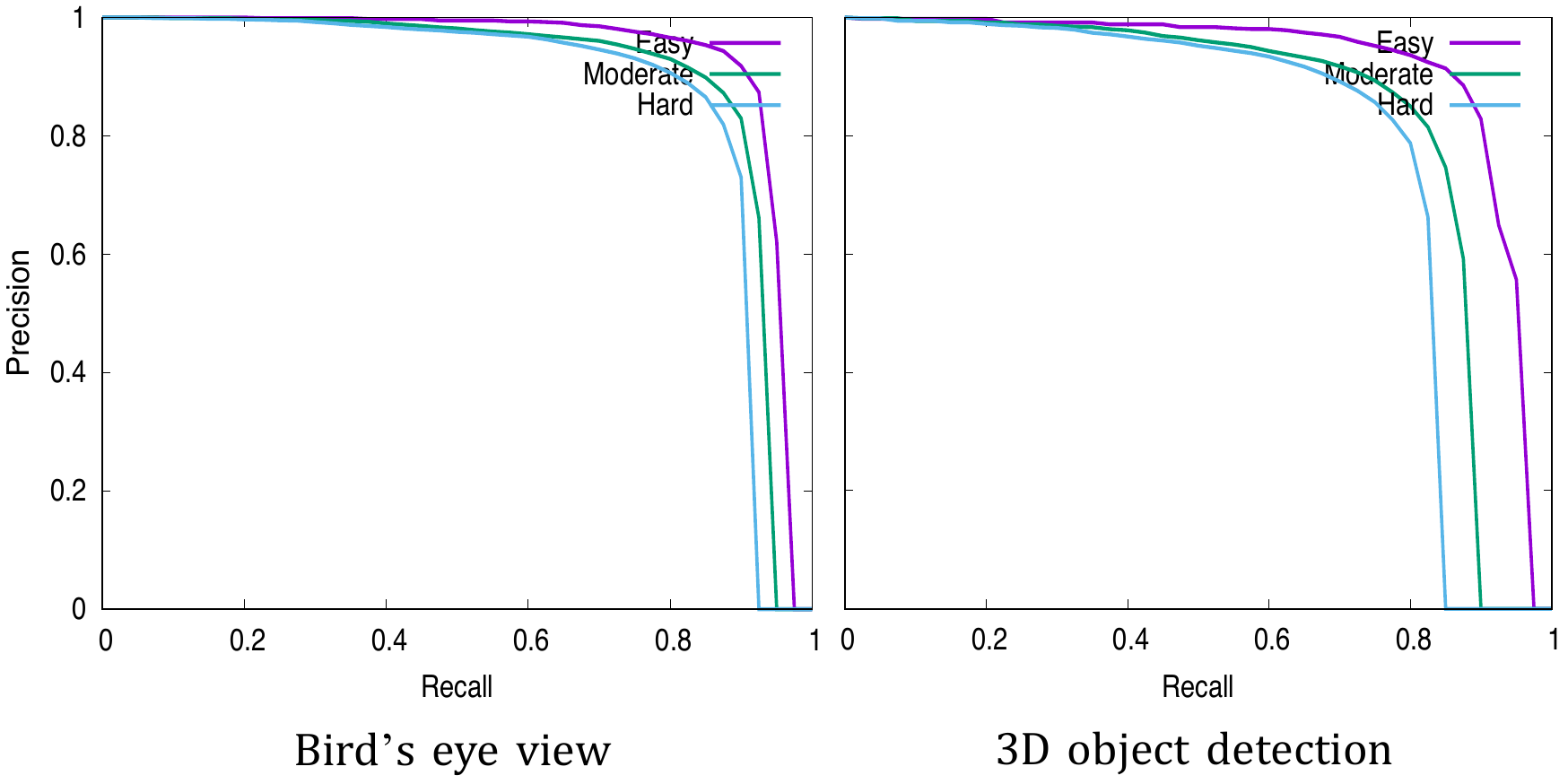}
   \caption{PR curves of \OursF on KITTI {\em test}.}
   \label{fig:kitti-pr-curves}
\end{center}
\end{figure}
\section{Backbone Networks}
\label{sec:backbone-networks}
We illustrate the structure of the backbone networks in Fig.~S~-~\ref{fig:backbone-network}. In this illustration, {\em Reg block} and {\em Subm block} mean the regular sparse convolutional block and the submanifold sparse convolutional block, respectively. The backbone networks are based on VoxelNet~\cite{voxelnet}.  It contains a stem layer and 4 stages. In the last three stages, Stage 1, 2, and 3, there a regular sparse convolutional block with stride as 2 for down-sampling. There are some detailed differences among different frameworks, as the following. 
\subsection{Architecture settings}
\noindent
\textbf{PV-RCNN and Voxel R-CNN.}
In the backbones of PV-RCNN~\cite{pvrcnn} and Voxel R-CNN~\cite{voxel-rcnn}, the channels for the stem and stages, \{$c_0$, $c_1$, $c_2$, $c_3$, $c_4$\}, are \{16, 16, 32, 64, 64\}. The numbers of {\em Subm blocks} in these stages, \{$n_1$, $n_2$, $n_3$, $n_4$\}, are \{1, 2, 2, 2\}. A {\em Reg} or {\em Subm block} is a conv-bn-relu layer, which includes a regular or submanifold convolution, a batch normalization layer~\cite{batchnorm}, and a ReLU activation.

\vspace{0.5em}
\noindent
\textbf{CenterPoint.}
In the backbone network of the CenterPoint~\cite{centerpoint} detector, the backbone network is larger. The channels for the stem and stages, \{$c_0$, $c_1$, $c_2$, $c_3$, $c_4$\}, equal to \{16, 16, 32, 64, 128\}. The numbers of repeated {\em Subm blocks} in these stages, \{$n_1$, $n_2$, $n_3$, $n_4$\}, are \{2, 2, 2, 2\}. Compared to that in the PV-RCNN~\cite{pvrcnn} and Voxel R-CNN~\cite{voxel-rcnn} detectors, the {\em Subm block} is more complicated in this backbone network. Except the stem, it contains two sequential conv-bn-relu layers, with a residual connection.

\begin{figure*}[t]
\begin{center}
   \includegraphics[width=\linewidth]{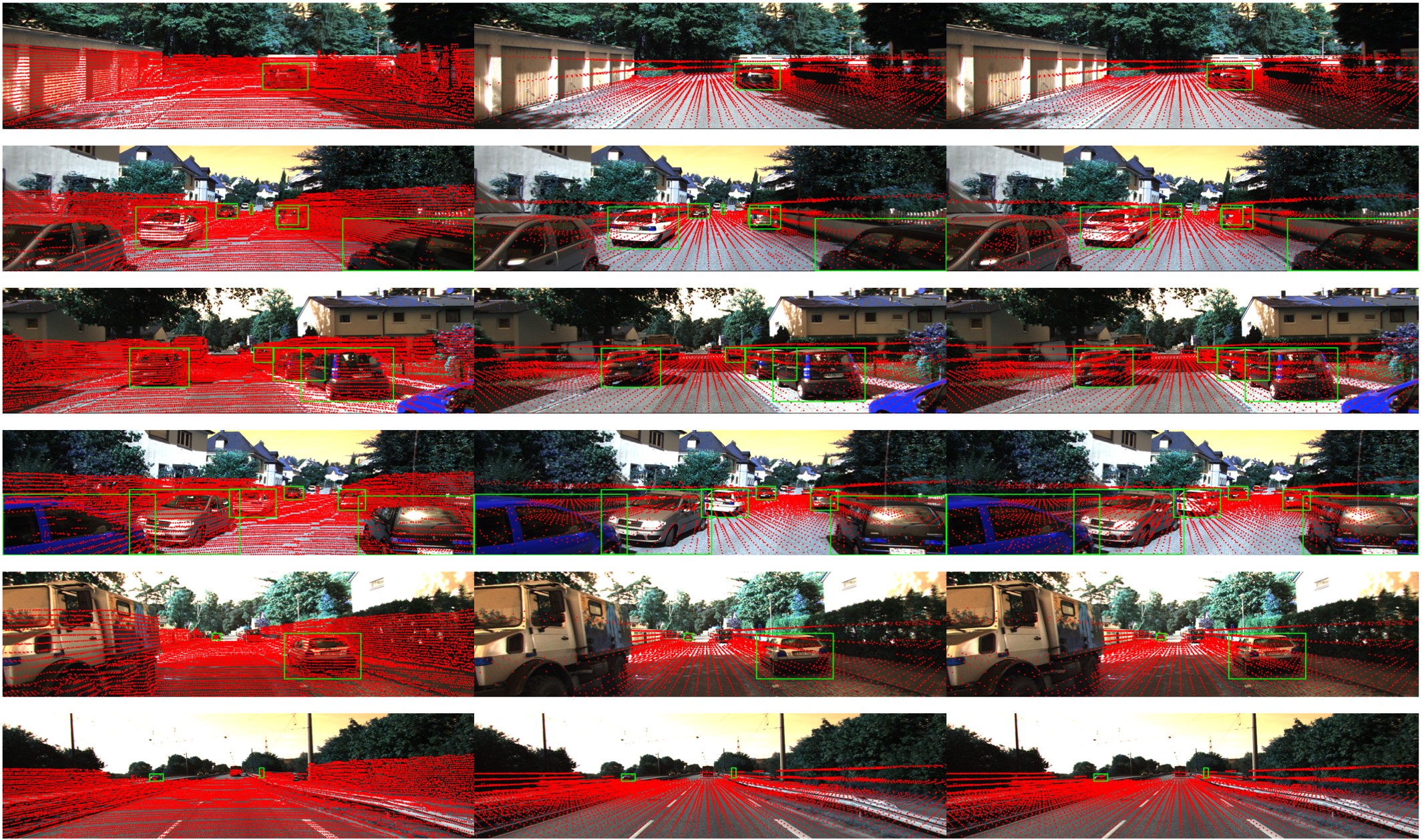}
   \caption{The illustration of visual comparisons between the plain and the focal sparse convolutional backbone networks.}
   \label{fig:illustration}
\end{center}
\end{figure*}

\subsection{Focal Sparse Convolution Usage}
In our approach, the above architecture-level settings are directly inherited from the original PV-RCNN~\cite{pvrcnn}, Voxel R-CNN~\cite{voxel-rcnn}, and CenterPoint~\cite{centernet} frameworks, without any adjustment, for a fair comparison. 
In the LIDAR-only task, we insert the {\em Focals Conv} in the last layer of Stage 1, 2, and 3. In the multi-modal task, we insert the {\em Focals Conv - F} only at the last layer of Stage 1. This relieves the efficiency and memory issues caused by the RGB feature extraction. Note that, in the CenterPoint~\cite{centerpoint} detectors, it is also used in the last {\em layer}, not the total {\em block}. In other words, although there are two conv-bn-relu layers in each Subm block in CenterPoint~\cite{centerpoint}, we only apply it as the last layer. For simplification, we do not double it as a block. 

\section{Additional Experiments}
\subsection{Results on Bird's Eye View on KITTI}
We report the accuracy for 3D object detection and Bird's Eye View (BEV) of \OursF upon Voxel R-CNN~\cite{voxel-rcnn} on the KITTI~\cite{kitti} dataset in Tab.~S~-~\ref{tab:kitti-val-r40}. The results are calculated by recall 40 positions with the IoU threshold of 0.7. It performs better than the strong Voxel R-CNN~\cite{voxel-rcnn} baseline on both AP$_{\textrm{BEV}}$ and AP$_{\textrm{3D}}$ in moderate and hard cases. 
We also provide the Prevision-Recall (PR) curves of \OursF on KITTI {\em test} split in Fig.~S~-~\ref{fig:kitti-pr-curves}.

\subsection{Objective Loss Weight}
\label{sec:ablations}
The training of the focal sparse convolutional networks involves the objective loss function. We implement it as a focal loss~\cite{focalloss} as in Eq~\eqref{eq:objloss}.
\begin{equation}
    L_{obj}=\frac{1}{|N|}\sum_{i\in \mathnormal{N}} -(1-\bar{p_{i}})^{\gamma}\mathrm{log}(\bar{p_{i}}).
    \label{eq:objloss}
\end{equation}
where $i\in N$ enumerates all sparse features in the current feature space. Following the original focal loss~\cite{focalloss}, we directly set $\gamma=2$ and it works well.
For notational convenience, we define $\bar{p_{i}}$ as follow
\begin{equation}
\bar{p_{i}}=\left\{
\begin{aligned}
& p_i, & \mathrm{if}\;\;\, y_i=1, \\
& 1 - p_i, & \mathrm{otherwise},
\end{aligned}
\right.
\label{eq:pi}
\end{equation}
where $p_i\in[0,1]$ is the estimated probability for the class with label $y_i=1$. It is the estimation that whether the feature $i$ contributes any foreground objects.

We analyze the loss weight for this objective loss in Tab.~S~-~\ref{tab:ablation-lossweight}. This ablation study is conducted upon the PV-RCNN~\cite{pvrcnn} detector on the KITTI~\cite{kitti} datasets. The results on AP$_{\textrm{3D}}$ with 40 recall positions are reported as the metric. We change the loss weight values from \{0.1, 0.5, 1.0, 2.0\}. It shows that too large or too small loss weight values degrade the results. Loss weights 0.5 and 1.0 present
competitive performance. We remain the 1.0 loss weight as a default setting for simplification.

\subsection{Improvements on the Waymo Open Dataset.}
To show our generalization capacity, we conduct further experiments on Waymo dataset. We use $\frac{1}{5}$ training data, following the default setting in the OpenPCdet codebase~\footnote{https://github.com/open-mmlab/OpenPCDet}. As shown in Tab.~S~-~\ref{tab:waymo}, \Ours also brings non-trivial improvements on the Waymo~\cite{waymo} dataset.

\subsection{Improvements on the nuScenes val split.}
Tab.~S~-~\ref{tab:improvements-over-centerpoint} presents the improvements over CenterPoint~\cite{centerpoint} on the nuScenes {\em val} split. The CenterPoint baseline in Tab.~S~-~\ref{tab:improvements-over-centerpoint} is re-implemented in the same settings to \Ours and \OursF. It shows that both \Ours and \OursF bring non-trivial improvements. Notably, \OursF improves the plain CenterPoint~\cite{centerpoint} by 4.9\% mAP on the nuScenes {\em val} split. We further apply some tricks for performance enhancement, {\em e.g.}, disabling ground-truth sampling in the last 4 epochs~\cite{pointaugmenting} and double-flip testing~\cite{centerpoint}.

\subsection{Accuracy loss on some categories after fusion.}
A surprising case is that the multi-modal fusion make the performance stay the same or worse on some popular categories, {\em e.g.}, Car, Ped, Bar (from {\em Focals Conv} to {\em Focals Conv - F} in Tab.~\textcolor{red}{11}). The improvements over the baseline are consistent on all categories. 
To analyze this special case, we conduct ablations on augmentations on CenterPoint and the nuScenes $\frac{1}{4}$ training set. We find {\em ground-truth sampling}~(GT Sampl.) is the keypoint. 
As in Tab.~A~-~\ref{tab:nuscenes-1/4-augs}, when GT Sampl. is used, the performance on some popular categories ({\em e.g., Car, Bus, Ped, Bar}) stays the same or worse. In contrast, when we disable GT Sampl. and apply all other transformations (flip, rotation, re-scaling, and translation), all categories are benefited from the fusion.
We suppose that this is from the image-level copy-paste in GT Sampl. When other objects are pasted onto images, popular objects inevitably have more chance to be covered by the pasted, which degrades the performance on these categories.

\subsection{Ablations on Voxel Size.}
We ablate the effects of different voxel sizes upon \OursF on the nuScenes {\em val} split in Tab.~S~-~\ref{tab:ablations-voxelsize}. We change the voxel sizes in {\em X} and {\em Y} axes from 0.05m to 0.15m, with the interval 0.025m. The overall mAP achieves the best performance at the voxel size (0.075, 0.075, 0.2)m.
However, the proper voxel sizes vary across different classes. This phenomenon deserves further analysis or a dynamic mechanism design in the future.

\section{Visualizations}
\label{sec:Visualizations}
We provide additional visual comparisons between the plain network and the focal sparse convolutional networks in Fig.~S~-~\ref{fig:illustration}. It shares the same settings to the Fig.~\textcolor{red}{2} in the paper. These visualizations are based on the PV-RCNN~\cite{pvrcnn} detectors and on the KITTI~\cite{kitti} dataset. In each visualization group, the top figure is the distribution of input voxels. The middle and the bottom figures are from the plain and the focal sparse convolutional networks, respectively. We project the coordinate centers of the output voxel features from the backbone networks onto the 2D image plane. The projection is based on the calibration matrices of KITTI~\cite{kitti}. 

\end{document}